\documentclass[conference]{IEEEtran}
\IEEEoverridecommandlockouts
\usepackage{cite}
\usepackage{amsmath,amssymb,amsfonts}
\usepackage{algorithmic}
\usepackage{graphicx}
\usepackage{subcaption}
\usepackage{textcomp}
\usepackage{xcolor}
\usepackage{dblfloatfix}
\usepackage[numbers]{natbib}
\usepackage[normalem]{ulem}

\def\BibTeX{{\rm B\kern-.05em{\sc i\kern-.025em b}\kern-.08em
    T\kern-.1667em\lower.7ex\hbox{E}\kern-.125emX}}
\begin{document}

\newcommand{\mahsa}[1]{\textcolor{teal}{#1}}
\newcommand{\mahsacm}[1]{\textcolor{blue}{#1}}

\title{LLM Inference Acceleration via \\
Efficient Operation Fusion
}

\author{\IEEEauthorblockN{Mahsa Salmani}
\IEEEauthorblockA{\textit{d-Matrix} \\
Santa Clara, CA, USA \\
msalmani@d-matrix.ai}
\and
\IEEEauthorblockN{Ilya Soloveychik}
\IEEEauthorblockA{\textit{d-Matrix} \\
Santa Clara, CA, USA \\
ilyas@d-matrix.ai}
}

\maketitle

\begin{abstract}
The rapid development of the Transformer-based Large Language Models (LLMs) in recent years has been closely linked to their ever-growing and already enormous sizes.
Many state-of-the-art language models contain hundreds of billions to trillions of parameters and require dedicated hardware resources for both training and inference. One of the key challenges inherent to the Transformer architecture is the requirement to support numerous non-linear transformations that involves normalization. For instance, each decoder block typically contains at least one Softmax operation and two Layernorms. 
The computation of the corresponding normalization scaling factors becomes a major bottleneck because it requires spatial collective operations. In other words, when it comes to the computation of denominators for Softmax and Layernorm, all vector elements must be aggregated into a single location, requiring significant communication. These collective operations slow down inference on Transformers by approximately 20\%, defeating the whole purpose of distributed in-memory compute.

In this work, we propose an extremely efficient technique that can completely hide the overhead caused by such collective operations. Note that each Softmax and Layernorm operation is typically followed by a large linear layer. Since non-linear and linear operations are performed on different hardware engines, they can be easily parallelized once the algebra allows such commutation. By leveraging the inherent properties of linear operations, we can defer the normalization of the preceding Softmax and Layernorm until after the linear layer is computed. Now we can compute the collective scaling factors concurrently with the matrix multiplication and completely hide the latency of the former behind the latter. Due to the algebraic equivalence, such parallelization preserves the numerical accuracy while significantly improving the hardware utilization and reducing the overall latency.


\end{abstract}

\begin{IEEEkeywords}
large language models, inference acceleration, operation fusion, collective operations, in-memory computing 
\end{IEEEkeywords}

\section{Introduction}
Transformer-based architectures \cite{b1}, particularly large language models (LLMs), have recently gained significant attention due to their impressive capabilities and performance across a wide range of tasks. These LLMs, such as GPT series \cite{brown} and LLama series \cite{touvron}, have demonstrated exceptional potential across a wide range of practical applications, including text generation, conversational processing, and knowledge-based question answering. However, as the size of LLMs grows, their hardware and memory requirements increase significantly, limiting their applicability and hindering their deployment in real-world scenarios. 

Various approaches targeting different aspects of compute optimization have been studied to address the computational and memory issues arising in such LLMs, including model compression techniques \cite{wang, zhu}, knowledge distillation \cite{ho}, pruning \cite{xia, chen}, development of lower precision formats \cite{rohani1, rohani2} for efficient quantization \cite{franter, jin}, approximations, and other software optimizations \cite{qin}. These methods are commonly employed to reduce both the computational demands and latency of inference in Transformers. 

Notably, most of such methods make the compute more efficient at the expense of model performance, as each compression technique introduces errors through approximations. Consequently, any approach that can enhance computational efficiency or reduce latency in high-dimensional computations while preserving the accuracy of the underlying system becomes of paramount importance. These advancements would be crucial in ensuring the practical deployment and scalability of large-scale LLMs, particularly in real-time applications where both speed and accuracy are critical.


One of the main challenges in executing Transformer-based LLMs on existing hardware lies in handling numerous non-linear functions that are not necessarily hardware-friendly. In particular, non-linear operations involving normalization, such as Layernorm and Softmax, require computation of the denominator, which necessitates data aggregation across multiple processing units. Such collectives sitting on the critical path often take much time to execute \cite{coll_commun, hoefler} and create numerous bottlenecks.

To address these challenges, various parallelization techniques have been extensively explored in the literature, aiming to ultimately enhance the computational efficiency and reduce the latency caused by these operations during LLM inference \cite{shoeybi, kim}. 
However, the existing parallelization approaches primarily target higher-level computations, such as data, model, or even pipeline parallelization.

In this work, we introduce a novel approach to enhance computational efficiency and reduce latency specifically at the Transformer-layer level. In particular, we present a latency reduction strategy for computing Layernorm and Softmax in Transformer-based LLMs. Our technique leverages an innovative operation fusion strategy that takes advantage of the inherent structure of the Transformer block architecture and the sequential ordering of layers within each block to streamline and accelerate the computation process. By targeting execution optimization at the granular level of Transformer layers, we ensure a balance between efficiency and performance without compromising model integrity. Importantly, our approach derives algebraically-equivalent equations, avoiding the need for quantization or approximations, thus preserving the model's accuracy.

\vspace{1em}
\noindent \textbf{Notation.}
The following notation is used in the article. Matrices are denoted by capital bold letters $\mathbf{M}$ and vectors by lower case bold $\mathbf{v}$. The operator product of matrices $\mathbf{A}$ and $\mathbf{B}$ of appropriate sizes is written as $\mathbf{A}\cdot\mathbf{B}$ or $\mathbf{A}\mathbf{B}$, while their element-wise product would be denoted by $\mathbf{A}\odot\mathbf{B}$. 
Given vector $\mathbf{m}$, we denote by $\mathbf{M} = \operatorname{Diag}(\mathbf{m})$ the diagonal matrix with elements of $\mathbf{m}$ on the main diagonal.

\section{Methodology}
\label{motiv}
One of the key challenges associated with the Transformer architecture is the presence of numerous non-linear transformations that involve normalization. Such normalization inherent in operations like Softmax and Layernorm is essential for ensuring the correct scaling of activations, preventing them from exploding or shrinking during training. The computation of such normalization scaling factors necessitates spatial collective operations. More specifically, the calculation of denominators of Softmax and Layernorm requires the aggregation of vector elements into a single location for proper processing. These collective operations are so time-consuming and introduce substantial delays, significantly slowing down the inference process. More importantly, such spatial aggregation defeats the entire purpose of distributed or in-memory compute that all hardware accelerators for LLMs are trying to achieve.

In this work, we propose an efficient technique that can completely eliminate the overhead caused by such collective operations in the Transformer architectures. The main idea is based on the observation that all the non-linearities at hand are followed by matrix multiplications. Being a linear operation matrix multiplication commutes with scaling, thus allowing the normalization of Softmax or Layernorm to be deferred and performed immediately after the multiplication. Leveraging this commutativity allows us to completely hide the computation of the denominators behind the corresponding matrix multiplications since these are performed on separate hardware components. The former is executed by the Single Instruction Multiple Data (SIMD) unit \cite{hughes}, while the latter is performed on a Digital In-Memory Compute (DIMC) unit \cite{mannocci, wang_dimc}. In other words, we fuse the respective non-linear operations with their subsequent matrix multiplications into single operations, thereby accelerating the overall computation. 

\begin{figure}[htbp]
    \centering
    \begin{subfigure}[b]{0.45\columnwidth}
        \centering
        \includegraphics[width=0.6\linewidth]{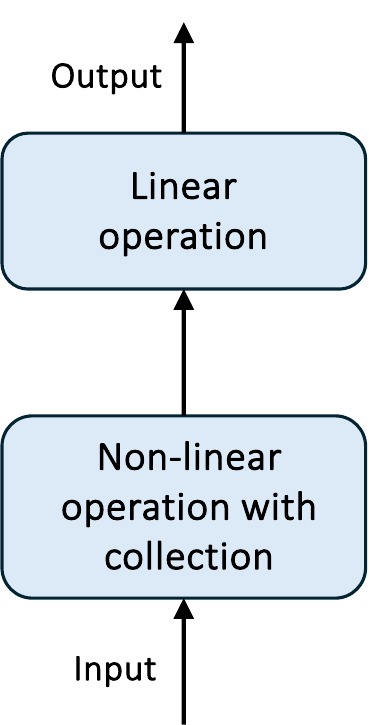} 
        \caption{}
        \label{fig_conven_op_arc}
    \end{subfigure}
    \hfill
    \begin{subfigure}[b]{0.53\columnwidth}
        \centering
        \includegraphics[width=\linewidth]{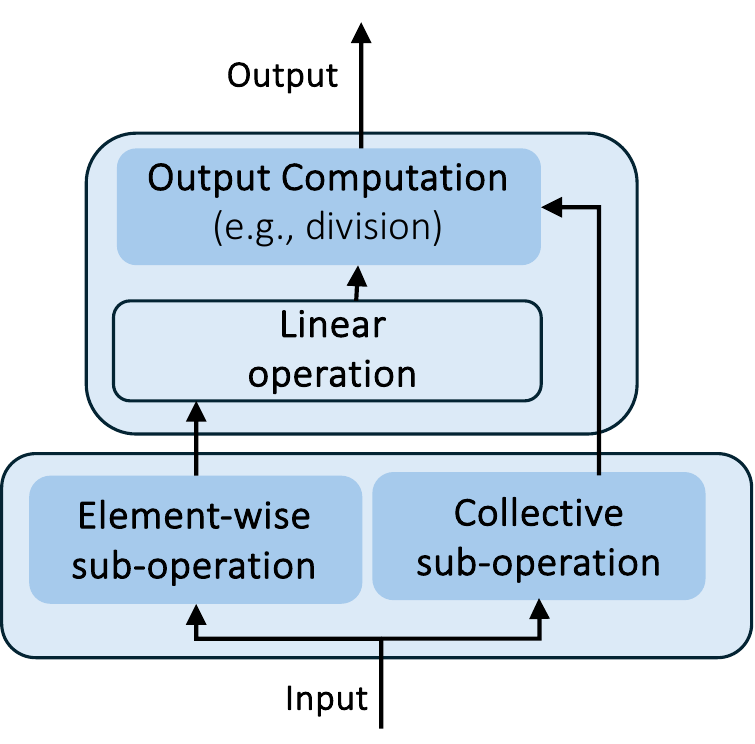} 
        \caption{}
        \label{fig_fused_op_arc}
    \end{subfigure}
    \caption{Conventional (a) vs. fused-operation (b) architecture.}
    \label{gen_archs}
\end{figure}

Fig.~\ref{gen_archs} compares the conventional architecture and the proposed fused architecture. The non-linear operation with collection in Fig.~\ref{fig_conven_op_arc} denotes either a Layernorm or a Softmax block, which can be decomposed into two sub-operations as discussed above: an element-wise sub-operation and a collective sub-operation, see Fig.~\ref{fig_fused_op_arc}. The element-wise sub-operation can be fused by the linear operation, which is the proceeding linear layer. By executing the aggregation in the collective sub-operation concurrently with the matrix multiplication between the element-wise sub-operation and the linear operation, the latency imposed by the collective operations can be effectively hidden, significantly accelerating the overall computation.

\begin{figure*}[t]
\centering
\centerline{\includegraphics[width=0.68\textwidth]{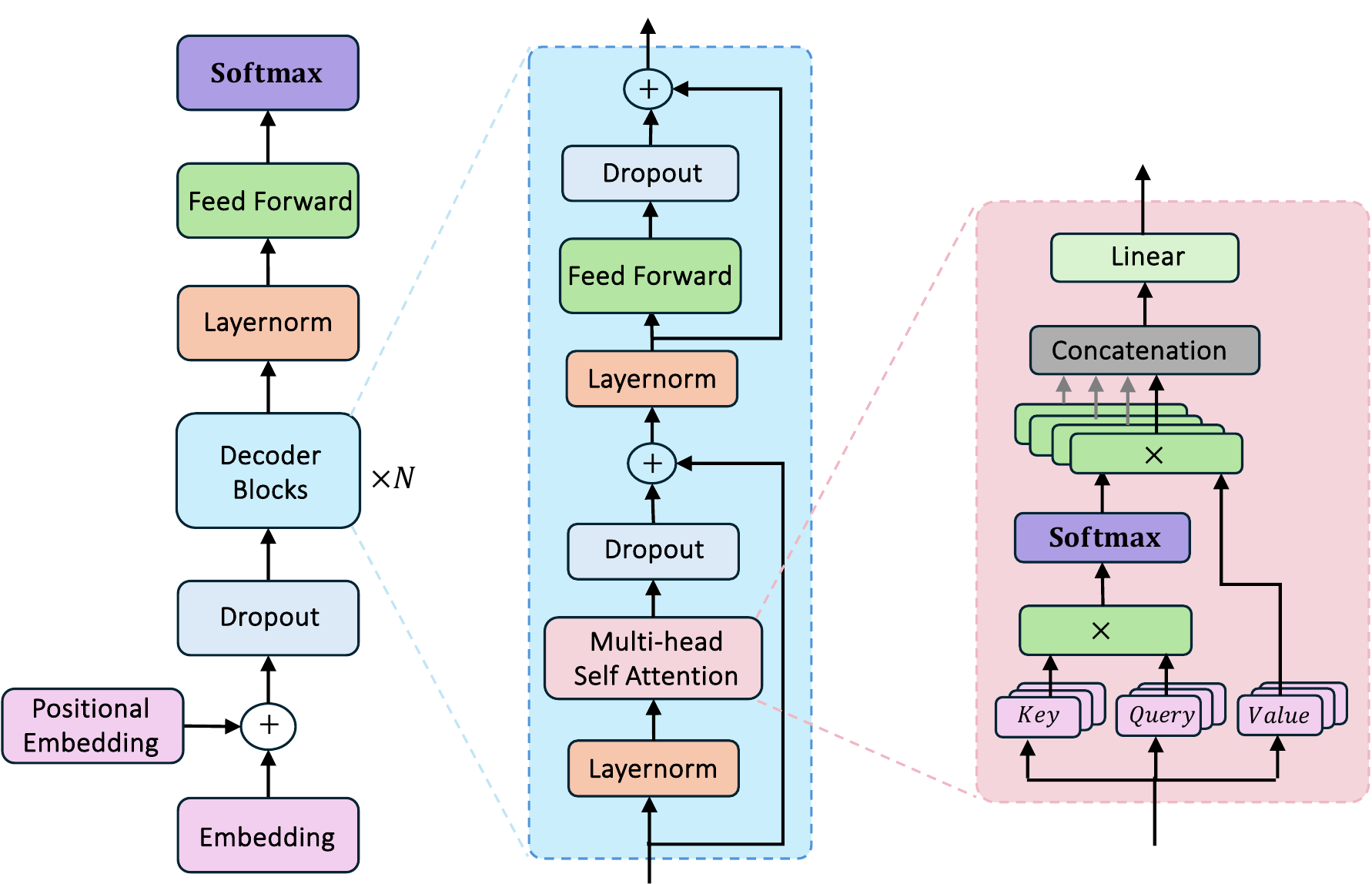}}
\caption{Transformer architecture.}
\label{total_arc}
\end{figure*}

Fig.~\ref{total_arc} demonstrates the typical architecture of a Transformer-based model. Each decoder block of such a model would contain one Softmax operation and two Layernorms, each followed by a matrix multiplication. Application of our fusion technique can reduce the total latency of inference on such LLMs by approximately 15-20\% for different hardware architectures.

Next, we explain how the fusion technique works in practice for both Layernorm and Softmax operations. We would like to emphasize that the fusion methodology proposed here guarantees that the fused operation produces exactly the same results as the original sub-graph due to their algebraic equivalence.

\subsection{Fused Layernorm}
Layernorm is one of the key components in modern LLMs \cite{xiong_laynorm}. It adaptively scales the input across the features for each data point and thus
helps stabilize the learning process.

Given the input vector $\mathbf{x} \in \mathbb{R}^{1\times n}$, the Layernorm first computes the mean and the variance of the elements as 
\begin{equation}
\label{eq_mean_var}
\bar{\mathbf{x}} = \tfrac{1}{n} \sum_1^{n} x_i, \quad \sigma^2(\mathbf{x}) = \tfrac{1}{n} {(\mathbf{x}-\bar{\mathbf{x}} 
\mathbf{1})(\mathbf{x}-\bar{\mathbf{x}}
\mathbf{1})^T},
\end{equation}
and after normalization, Layernorm applies trainable scale factors $\boldsymbol{\gamma}$ and bias $\boldsymbol{\beta}$ to allow the model to adjust the normalized values as follows
\begin{equation}
\label{eq_norm}
\mathbf{y} =\frac{\mathbf{x}-\bar{\mathbf{x}} \mathbf{1}}{{\sqrt{\sigma^2+\epsilon}}} \odot \boldsymbol{\gamma} + \boldsymbol{\beta}.
\end{equation}
If $\boldsymbol{\Gamma} \in \mathbb{R}^{n \times n}$ denotes the diagonal matrix whose diagonal elements are 
the components of the vector $\boldsymbol{\gamma}$, i.e., $\boldsymbol{\Gamma}= \operatorname{Diag}(\boldsymbol{\gamma})$, and $\mathbf{E} \in \mathbb{R}^{n \times n}$ denotes a matrix in which all elements are equal to 1, Eq.~\ref{eq_norm} can be expressed in matrix form as
\begin{equation}
\label{eq_fused_layernorm_final}
\mathbf{y} = \frac{\mathbf{x}-\bar{\mathbf{x}} \mathbf{1}}{\sqrt{\sigma^2+\epsilon}} \odot \boldsymbol{\gamma} + \boldsymbol{\beta} =  \frac{\mathbf{x}} {\sqrt{\sigma^2+\epsilon}} \left ( \mathbf{I} - \tfrac{1}{n} \mathbf{E} \right )  \boldsymbol{\Gamma} + \boldsymbol{\beta} .
\end{equation}
Note that $\mathbf{I}$, $\mathbf{E}$, and $\boldsymbol{\Gamma}$ are all static matrices, and hence $\left ( \mathbf{I} - \tfrac{1}{n} \mathbf{E} \right ) \boldsymbol{\Gamma}$ in Eq.~\ref{eq_fused_layernorm_final} can be computed at the compile time with no impact on the inference latency.

Finally, the output of the Layernorm is directly fed into the following linear layer. Namely, if $\mathbf{F} \in \mathbb{R}^{n \times m}$, the result of the matrix multiplication, $\mathbf{yF}$, can be written as 
\begin{equation}
\mathbf{y} \mathbf{F} =  \underbrace{\left( \frac{\mathbf{x}} {\sqrt{\sigma^2+\epsilon}} \Bigl(\mathbf{I} - \tfrac{1}{n} \mathbf{E} \Bigr)  \boldsymbol{\Gamma} + \boldsymbol{\beta} \right )}_{\text{conventional Layernorm}} \mathbf{F}.
\end{equation}

The standard Transformer architecture prescribes to first complete the Layernorm and then perform the matrix multiplication, following the architecture in Fig.~\ref{fig_conven_op_arc}. 

\begin{figure}[h]
\centerline{\includegraphics[width=0.4\textwidth]{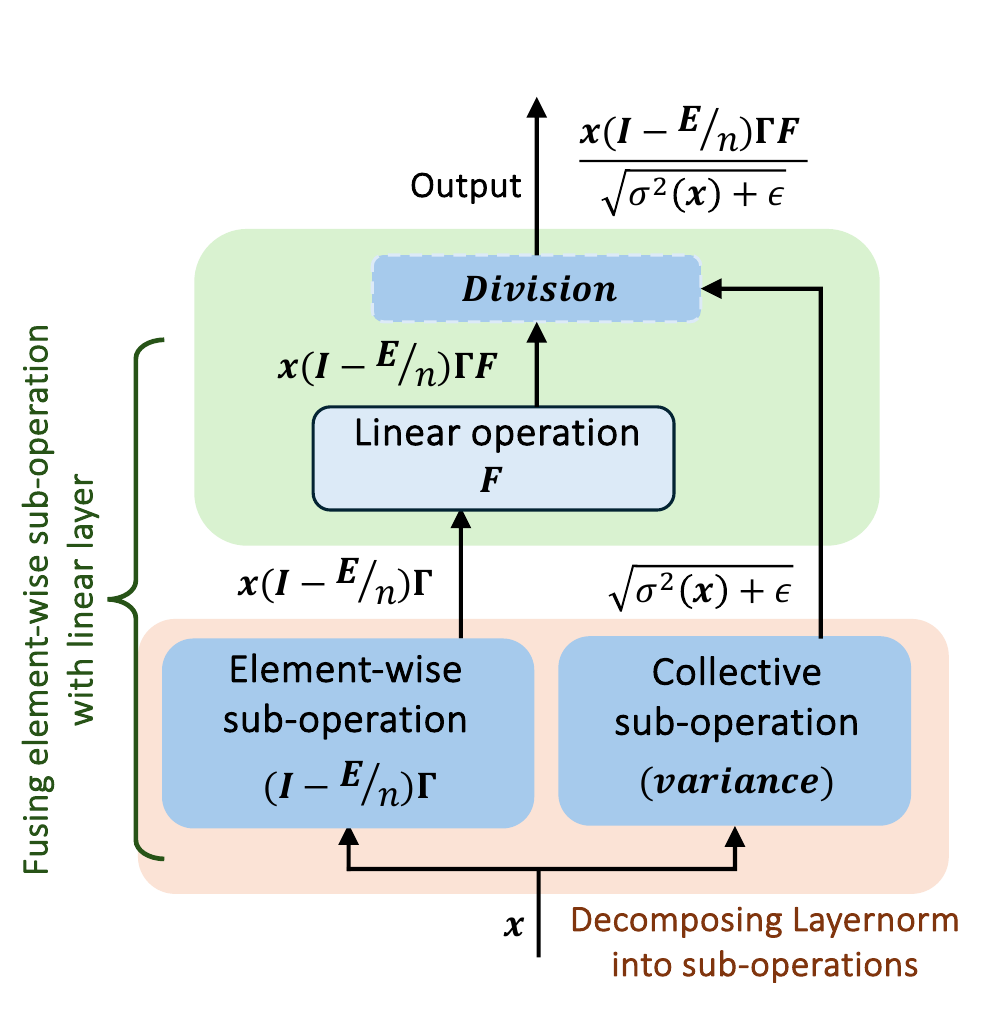}}
\caption{Architecture of fused Layernorm.}
\label{fused_norm}
\end{figure}

Our proposed fusion methodology, instead, suggests following an operation decomposition as below: 
\begin{equation}
\label{eq_fused_layernorm}
\mathbf{y} \mathbf{F} = \underbrace{\frac{1}{\sqrt{\sigma^2+\epsilon}}}_{\text{collective sub-op}} \times \underbrace{\Bigl( \mathbf{x} \big(\mathbf{I} - \frac{1}{n} \mathbf{E} \big)  \boldsymbol{\Gamma}\mathbf{F} \Bigr)}_{\text{fused Layernorm}}   + \boldsymbol{\beta}  \mathbf{F}.
\end{equation}
In  Eq.~\ref{eq_fused_layernorm}, the Layernorm computation is first decomposed into an element-wise computation, i.e., $\mathbf{x} \big(\mathbf{I} - \frac{1}{n} \mathbf{E} \big) \boldsymbol{\Gamma}$, and a denominator with aggregative operations, i.e., $\sqrt{\sigma^2+\epsilon}$. Then, exploiting the commutative property, while the linear layer is executed as $\mathbf{x} \big(\mathbf{I} - \frac{1}{n} \mathbf{E} \big) \boldsymbol{\Gamma} \mathbf{F}$ in the linear processing engine (e.g., DIMC), the aggregation required for the denominator is concurrently performed in the non-linear processing engine (e.g., SIMD). Fig.~\ref{fused_norm} illustrates the architecture of the fused Layernorm, highlighting the operation decomposition and fusion process. Note that in this architecture, $\boldsymbol{\beta}$ is omitted for simplicity but it can be computed as part of the linear operation, if required.

We would like to note that in the Transformer block, as shown in Fig.~\ref{total_arc}, there are two different architectural configurations with respect to Layernorm and the residual connection. In this work, we describe the architecture where the residual connection is branched before the Layernorm step. However, our proposed fused Layernorm architecture can be easily applied to the configuration where the residual connection is branched immediately after Layernorm.


\subsubsection*{RMSNorm in Llama Models}

The Llama family of models exploit a different type of normalization instead of Layernorm---the so-called RMSNorm \cite{{zhang_rms}}. RMSNorm assumes the mean of the activation vector is zero and thus reads as 
\begin{equation}
\label{rms_norm}
\operatorname{RMSNorm}(\mathbf{x}) = \frac{\mathbf{x}}{\sqrt{\tfrac{1}{n} \mathbf{x} \mathbf{x}^T}} \odot \boldsymbol{\gamma}.
\end{equation}
Additionally, the subsequent linear layer in Llama’s Multi-Layer Perceptron (MLP) differs from the standard Transformer MLPs by incorporating an up-projection, a gating mechanism (e.g., SwiGLU), and a down-projection. Specifically, in Llama’s MLP, the input is first projected into a higher-dimensional space (up-projection), then modulated by a gating function that selectively adjusts activations, and finally reduced back to the original dimensionality (down-projection). 

Despite these architectural distinctions, our operation fusion approach integrates seamlessly with Llama models' layers. By leveraging algebraic equivalences, our method fuses computations in the up-projection and gating stages, preserving the exact accuracy of the original layer while reducing latency through optimized intermediate operations.

\subsection{Fused Softmax}
Similarly to Layernorm, the Softmax in Transformers is typically followed by a matrix multiplication. If $\mathbf{x}, \mathbf{y}\in \mathbb{R}^{1\times n}$ denote the input and output vectors of a Softmax\footnote{In practice, the computation involves computing the element-wise maximum, $\operatorname{max}_i(\mathbf{x})$, and then applying Softmax function to $\mathbf{x}-\operatorname{max}_i(\mathbf{x})$. Note that the fusion methodology applies in this case without alterations.}, respectively,
the output of the following matrix multiplication reads as
\begin{equation}
\mathbf{y} \mathbf{V} = \underbrace{\Bigl( \frac{[e^{x_1}~e^{x_2}~\cdots~e^{x_n}]}{\sum_i e^{x_i}}\Bigr)}_{\text{conventional Softmax}}  \mathbf{V},
\end{equation}
where $\mathbf{V} \in \mathbb{R}^{n \times m}$ denotes the Values matrix.

\begin{figure}[h]
\centering
\centerline{\includegraphics[width=0.4\textwidth]{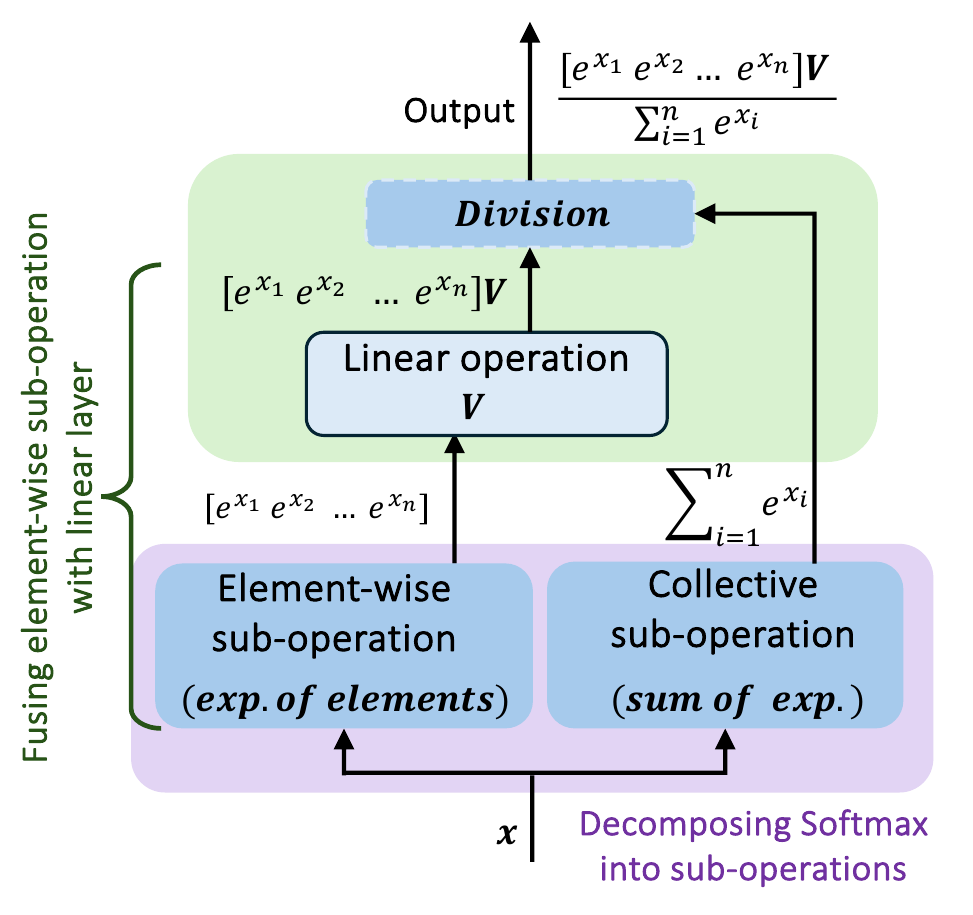}}
\caption{Architecture of fused Softmax.}
\label{fig_fused_softmax}
\end{figure}

Analogously to the treatment of Layernorms described above, the fusion methodology applies here as well. We can first compute the exponential numerators, use them for the matrix operation and then apply normalization as illustrated in Fig.~\ref{fig_fused_softmax}.


Note that unlike the Layernorm case, $\mathbf{V}$ is now a matrix of activations and not a static weight matrix. Despite that, the fusion algorithm is still perfectly applicable here because the matrix multiplication retains its linear nature.

\begin{equation}
\label{eq_fused_softmax}
\mathbf{y} \mathbf{V} = \underbrace{\frac{1}{\sum_i e^{x_i}}}_{\text{collective sub-operation}}  \underbrace{\Bigl([e^{x_1}~e^{x_2}~\cdots~e^{x_n}]\mathbf{V} \Bigr)}_{\text{fused Softmax}} .
\end{equation}

\vspace{-5pt}

\section{Experimental Results}
The proposed technique is based on an algebraically equivalent reformulation of the involved operations, ensuring that it maintains complete accuracy without any degradation. Therefore, our methodology can be easily applied to any architecture that involves a Layernorm or Softmax followed by a linear operation, in particular it applies to any Transformer-based LLM.
While the key advantage of this technique lies in its ability to significantly reduce the latency, its overall effectiveness largely depends on the underlying hardware architecture and the implementation of computations and collective operations. Key factors such as the throughput and capabilities of different computing units handling linear and non-linear operations play a crucial role in determining the performance gain achieved with the proposed method. 

In order to provide an experimental analysis, we implemented the proposed approach on Corsair \cite{corsair}, our recently launched AI accelerator as described in Section~\ref{motiv}. The experimental results demonstrate that we get a latency reduction of 20\% on most state-of-the-art LLMs such as Llama2 and LLama3. These findings highlight the effectiveness of the fusion strategy in optimizing Transformer-based computations, further validating its potential for accelerating Artificial Intelligence (AI) workloads on advanced hardware platforms. Due to space constraints, we defer the rigorous analysis of computational time gains to our next publication.

\section{Conclusion}
In this work, we introduced an operation fusion technique designed to enhance the computational efficiency and reduce the latency of Transformer-based large language models. By decomposing and fusing specific non-linear and linear operations, i.e., those that involve normalization, such as Layernorm and Softmax, with their succeeding linear transformations, we achieved mathematically equivalent results without compromising model accuracy. Deployment of the proposed approach on modern AI accelerators demonstrates a 20\% reduction of inference latency. These findings underscore the potential of operation fusion strategies to address the computational challenges posed by large scale Transformer-based models, paving the way for more efficient and scalable implementations in real-world applications. Future work will explore extending this methodology to other architectural components and further optimizing hardware-software co-design.



\vspace{12pt}
\color{red}

\end{document}